\documentclass[11pt]{article}

\usepackage{amssymb,amsmath,amsthm,float,graphicx,color,amsmath,cases,graphics,verbatim,mathrsfs}

\usepackage[margin=1in]{geometry}

\definecolor{lightgray}{rgb}{0.95,0.95,0.95}

\newtheorem{theorem}{Theorem}
\newtheorem{lemma}{Lemma}

\newtheorem{assumption}{Assumption}

\def\[{\lf[} \def\]{\ri]}  \def\br{\right} \def\lf{\left} \def\ri{\right}

\def\old#1{}
\def\pf{{\smallskip\noindent\it Proof.  }}
\def\qed{{\hfill $\blacksquare$}}
\def\to{\rightarrow}

\def\E{{\bf E}} 
 
\def\argmin{{\rm argmin}}

\def\ek{ \mathcal{E}^t}

\def\a{ }

\def\cH{\mathcal{H}}
\usepackage{multirow}

\author{Mengdi Wang\\ \small Department of Operations Research and Financial Engineering, Princeton University, Princeton, NJ\\  \small 
email: mengdiw@princeton.edu}
\title{Primal-Dual $\pi$ Learning: Sample Complexity and Sublinear Run Time for Ergodic Markov Decision Problems}

\def\carS{|\cS|}

\def\carA{|\cA|}

\def\cA{\mathcal{A}}

\def\cO{\mathcal{O}}

\def\cF{\mathcal{F}}
\def\cM{\mathcal{M}}
\def\cS{\mathcal{S}}
\def\cH{\mathcal{H}}

\def \gk{\mathcal{G}^t}
   \def\cE{\mathcal{E}}

\usepackage{algpseudocode,algorithm}

\def\S{|\cS|}
\def\A{|\cA|}
\def\suma{\sum_{a\in \cA}}
\def\sumi{\sum_{i\in\cS}} 
\def\sumj {\sum_{j\in\cS}} 
\def\cF{\mathcal{F}}
\def\cP{\mathcal{P}}

 \def\bh{\mathbf{h}}
 \def\g{\Delta}
 \def\sumia{\sumi\suma}
\def\KL{D_{KL}}
\def\tO{\tilde\cO}
\def\cU{\mathcal{U}}
\def\cG{\mathcal{G}}
\def\tmix{t^*_{mix}}

\begin{document}

\maketitle

\begin{abstract}
Consider the problem of approximating the optimal policy of a Markov decision process (MDP) by sampling state transitions. In contrast to existing reinforcement learning methods that are based on successive approximations to the nonlinear Bellman equation, we propose a Primal-Dual $\pi$ Learning method in light of the linear duality between the value and policy.  The $\pi$ learning method is model-free and makes primal-dual updates to the policy and value vectors as new data are revealed. 
For infinite-horizon undiscounted Markov decision process with finite state space $\cS$ and finite action space $\cA$, the $\pi$ learning method finds an $\epsilon$-optimal policy using the following number of sample transitions 
$$  \tO\lf( \frac{(\tau\cdot t^*_{mix})^2 \S \A }{\epsilon^2} \ri),$$
where $t^*_{mix}$ is an upper bound of mixing times across all policies and $\tau$ is a parameter characterizing the range of stationary distributions across policies. 
The $\pi$ learning method also applies to the computational problem of MDP where the transition probabilities and rewards are explicitly given as the input. In the case where each state transition can be sampled in $\tO(1)$ time, the $\pi$ learning method gives a sublinear-time algorithm for solving the averaged-reward MDP. \end{abstract}

\noindent{\bf Keywords:}
    Markov decision process, reinforcement learning, sample complexity, run-time complexity, duality, primal-dual method, mixing time

\def\bp{\mathbf{p}}
\def\bpi{\mathbf{\mu}}
\def\bmu{\mathbf{\pi}}
\def\bv{\mathbf{v}}
\def\br{\mathbf{r}}
\def\be{\mathbf{1}}
\def\bq{\mathbf{q}}
\def\bx{\mathbf{x}}

\section{Introduction}

Consider the reinforcement learning problem in which a planner makes decisions in an unknown (sometimes stochastic) dynamic environment with the goal of maximizing the reward collected in this process. This can be modeled as a Markov decision process (MDP). MDP refers to a controlled random walk in which the planner chooses one from a number of actions at each state of the random walk and moves to another state according to some transition probability distribution. In the context of reinforcement learning, one wants to learn the optimal decision rule by using an algorithmic trial-and-error approach, without explicitly knowing the transition probabilities. 


We focus on the infinite-horizon Average-reward Markov Decision Problem (AMDP) in which one aims to make an infinite sequence of decisions and optimize the average-per-time-step reward. An instance of the AMDP can be described by a tuple $\mathcal{M}=(\cS,\cA,\cP, \br )$, where $\cS$ is a finite state space of size $\S$, $\cA$ is a finite action space of size $\A$, $\cP$ is the collection of state-to-state transition probabilities $\cP=\{ p_{ij}(a) \mid  i,j\in\cS,a\in\cA\}$, $\br$ is the collection of state-transitional rewards $\br=\{r_{ij}(a) \mid  i,j\in\cS,a\in\cA \}$ where $r_{ij}(a) \in [0, 1]$. We also denote by $\br_a$ the vector of expected state-transition rewards under action $a$, where $\br_{a,i} = \sumj p_{ij}(a)r_{ij}(a)$.
Suppose that the decision process is in state $i$, if action $a$ is selected, the process moves to a next state $j$ with probability $p_{ij}(a)$ and generates a reward $r_{ij}(a) $.  

We want to find a stationary policy that specifies which action to choose at each state (regardless of the time step). A stationary and randomized policy can be represented by a collection of probability distributions $\bmu = \{\bmu_i\}_{i\in\cS} $, where $\bmu_{i}:\cA\mapsto [0,1] $ is a vector of probability distribution over actions.  
We denote by $P^{\bmu}$ the transition probability matrix of the AMDP under a fixed policy $\pi$, where $P^{\bmu}_{ij} = \suma \bmu_{i}(a) p_{ij}(a)$ for all $i,j\in\cS$.
The objective of the AMDP is to find an optimal policy $\bmu^*$ such that the infinite-horizon average reward is maximized:
$$\max_{\bmu} \lim_{T\to\infty}  \E^{\bmu}\[{\frac1T} \sum^{T}_{t=1}    r_{i_{t} i_{t+1}}(a_t)\],$$
where $\{i_0, a_0, i_1,a_1, \ldots,i_t,a_t,\ldots\}$ are state-action transitions generated by the Markov decision process under the fixed policy $\bmu$, and the expectation $ \E^{\bmu}\[\cdot\]$ is taken over the entire trajectory. 

Let us emphasize our focus on the {\it undiscounted average-reward} MDP. This is contrary to the majority of existing literatures that focus on the discounted cumulative reward problems, i.e., $\max_{\bmu} \E^{\bmu}\[ \sum^{\infty}_{t=1}   \gamma^t r_{i_{t} i_{t+1}}(a_t)\]$ where $\gamma\in(0,1)$ is a pre-specified discount factor. The discount factor $\gamma$ is imposed artificially for analytical purposes. It ensures contractive properties of the Bellman operator and geometric convergence of value and policy iterations. It also plays an important role in the sample and run-time complexity analysis for MDP algorithms and reinforcement learning methods. However, discounted MDP are indeed approximations to infinite-horizon undiscounted MDPs . In this paper, we attempt to obsolete the discount factor.  Instead of assuming that future rewards are discounted, we focus on the undiscounted MDP that satisfies certain fast mixing  property and stationary properties. The lack of a discount factor significantly complicates our analysis.

\def\SO{$\mathcal{SO}$}

Let us focus on sampling-based methods for the AMDP. Suppose that $\mathcal{M}=(\cS,\cA,\cP, \br )$, is not explicitly given. Instead, it is possible to interact with the real-time decision process (or a simulated process) by trying different controls and observing states transitions and rewards. In particular, suppose that we are given a Sampling Oracle (\SO), which takes a state-action pair $(i,a)$ as input and outputs a random future state $j$ and reward $r_{ij}(a)$ with probability $p_{ij}(a)$. Such a \SO\ is known as the generative model in the literatures of reinforcement learning \cite{kearns1999finite, kearns2002sparse}.

In this paper, we propose a model-free policy learning method for solving the AMDP, which we refer to as Primal-Dual $\pi$ Learning ({\it $\pi$ learning} for short). It is motivated by a recently developed randomized primal-dual method for solving the discounted MDP \cite{wang2017randomized}. The $\pi$ learning method maintains a randomized policy for controlling the MDP and dynamically updates the policy and an auxiliary value vector as new observations are revealed. The method is based on a primal-dual iteration which is crafted to take advantage of the linear algebraic structures of the nonlinear Bellman equation. The $\pi$ learning method is remarkably computational efficient - it uses $\cO(\S\A)$ space and $\tO(1)$ arithmetic operations per update.\footnote{We use $\cO(1)$ to hide constant factors and use $\tO(1)$ to hide polylog factors of $\S,\A,\epsilon$.}
It is {\it model-free} in the sense that it directly updates the policy and value vectors without estimating the transition probabilities of the MDP model. 
We show that the $\pi$ learning method finds an $\epsilon$-optimal policy with probability $1-\delta$ using the following sample complexity (number of queries to the \SO):
$$  \tO\lf( \frac{(\tau\cdot t^*_{mix})^2\S \A }{\epsilon^2}\log \lf(\frac1{\delta}\ri) \ri),$$
where $\tau$ is parameter that characterizes the range of stationary distributions across policies, and $t^*_{mix}$ is an uniform upper bound of the mixing times of the Markov decision process under any stationary policy. 
This sample complexity is optimal in its dependence on $\frac{\S\A}{\epsilon^2}$.

When the MDP model $\mathcal{M}=(\cS,\cA,\cP, \br )$ is explicitly given, the proposed $\pi$ learning method can be used as a randomized algorithm to compute an $\epsilon$-optimal policy. Given $\mathcal{M}=(\cS,\cA,\cP, \br )$ as the input, one can implement \SO\ using binary-tree data structures using $\cO(\S^2\A)$ preprocessing time, such that each query to the \SO\ takes $\cO(1)$ time \cite{wang2017randomized}. In this setting, the $\pi$-learning method outputs an $\epsilon$-optimal policy with probability $1-\delta$ in run time $  \tO\lf( \frac{(\tau\cdot t^*_{mix})^2\S \A }{\epsilon^2}\log \lf(\frac1{\delta}\ri) \ri).$ This is a sublinear run time in comparison with the input size $\cO(\S^2\A)$, as long as $\epsilon\ll \sqrt{S}. $

To the author's best knowledge, 
this is the first model-free learning method for infinite-horizon average-reward MDP problems that is based on a primal-dual iteration. Our sample complexity result is a first result that characterizes the role of the mixing time and range of stationary distributions, without assuming any discount factor or finite horizon.
We also provide the first sublinear run-time result for approximately solving AMDP using randomization. 


\paragraph{Outline} Section 2 surveys existing model-free learning methods for MDP and their sample and run-time complexity guarantees. Section 3 states the main assumptions on the ergodic Markov decision processes, the Bellman equation and its linear programming formulations. Section 4 develops the Primal-Dual $\pi$ Learning method from a saddle point formulation of the Bellman equation. Section 5 establishes the convergence analysis and sample complexity of exploration of the Primal-Dual $\pi$ Learning method. 
Section 6 gives a summary.

\paragraph{Notations} All vectors are considered as column vectors. For a vector $\bx\in\Re^n$, we denote by $x_i$ or $x(i)$ its $i$-th component, denote by $\bx^{\top}$ its transpose, and denote by $\|\bx\| =\sqrt{\bx^{\top}\bx}$ its Euclidean norm. 
We denote by $\be = (1,\ldots,1)^{\top}$ the vector with all entries equaling 1, and we denote by $\be_i$ the vector with its $i$-th entry equaling $1$ and other entries equaling $0$.
For a positive number $x$, we denote by $\log x$ the natural logarithm of $x$.
For two probability distributions $p,q$ over a finite set $X$, we denote by $\KL(p||q) $ their Kullback-Leibler divergence, i.e., $\KL(p||q) = \sum_{x\in X} p(x) \log \frac{p(x)}{q(x)}$.

\section{Related Literatures}

There are two major notions of complexity for MDP: the run-time complexity and the sample complexity. The run-time complexity is critical to the computational problem where the MDP model is fully specified. It is measured by the total number of arithmetic operations performed by an algorithm. The sample complexity is critical to the reinforcement learning problem where the MDP model is unknown but a sampling oracle (\SO) is given. It is measured by the total number of queries to \SO\ made by an algorithm. Most existing literatures focus on either one of the two notions. They were considered as disjoint topics for years of research. 

The computational complexity of MDP has been studied mainly in the setting where the MDP model is fully specified as the input. Three major deterministic approaches are the value iteration method \cite{bellman1957dynamic, tseng1990solving, littman1995complexity} , the policy iteration method \cite{howard1960dynamic,mansour1999complexity,ye2011simplex,scherrer2013improved}, and linear programming methods \cite{d1963probabilistic, de1960problemes, ye2011simplex,scherrer2013improved,ye2005new}. These deterministic methods inevitably require solving large linear systems.
 In order to compute the optimal policy exactly or to find an $\epsilon$-optimal policy in $\tO\lf(\hbox{poly}(\S\A)\log(\frac1\epsilon) \ri)$ time, these methods all require linear or superlinear time, i.e., the number of arithmetic operations needed is at least linear in the input size $\cO(\S^2\A)$. 
For more detailed surveys on the exact solution methods for MDP, we refer the readers to the textbooks \cite{bertsekas1995dynamic, bertsekas1995neuro, puterman2014markov, bertsekas2013abstract} and the references therein.  

Randomized versions of the classical methods have proved to achieve faster run time when $\S,\A$ are very large. Examples include the randomized primal-dual method by \cite{wang2017randomized} and the variance-reduced randomized value iteration methods by \cite{soda2017}; both apply to the discounted MDP. These methods involve simulating the Markov decision processes and making randomized updates. As long as the input is given in suitable data structures that enable $\cO(1)$-time sampling, these results suggest that it is possible to compute an approximate policy for the discounted MDP in sublinear time $\tO(\frac{\S\A}{\epsilon^2})$ (ignoring other parameters).
On the other hand, \cite{chen2017lower} recently showed that the run-time complexity for any randomized algorithm is $\Omega(\S^2\A)$ for the discounted MDP. In the case where each transition can be sampled in $\tO(1)$ time, \cite{chen2017lower} showed that any randomized algorithm needs $\Omega(\frac{\S\A}{\epsilon})$ run time to produce an $\epsilon$-optimal policy with high probability. To the author's best knowledge, existing results on randomized methods only apply to the discounted MDP. It remains unclear how to use randomized algorithms to efficiently approximate the optimal average-reward policy.

The sample complexity of MDP has been studied mainly in the setting of reinforcement learning. In this paper, we are given a \SO\ that generate state transitions from any specified by state-action pair. This is known as the generative model in reinforcement learning, which was introduced and studied in \cite{kearns1999finite, kearns2002sparse}. In this setting, the sample complexity of the MDP is the number of queries to the \SO\ in order to find an $\epsilon$-optimal policy (or $\epsilon$-optimal value in some literatures) with high probability. 
One of the earliest reinforcement learning method is Q-learning, which are essentially sampling-based variants of value iteration. For infinite-horizon discounted MDP, \cite{kearns1999finite} proved that phased Q-learning takes $\tO(\frac{\S\A}{\epsilon^2})$ sample transitions to compute an $\epsilon$-optimal policy, where the dependence on $\gamma$ is left unspecified. 
Azar, Munos and Kappen \cite{azar2012sample} considered a model-based value iteration method for the discounted MDP and showed that it takes  $\tO\lf(\frac{\carS \carA }{(1-\gamma)^3\epsilon^2} \ri)$ samples to compute an $\epsilon$-optimal value vector (not an $\epsilon$-optimal policy). It also provided a matching sample complexity lower bound for estimating the value vector. It does not give explicit run-time complexity analysis. 

We summarize existing model-free sampling-based methods for MDP and their complexity results in Table 1. Note that the settings and assumptions in this works vary from one to another.
We also note that there is a large body of works on the {\it sample complexity of exploration} for reinforcement learning, which is the number of suboptimal time steps an algorithm performs on a single infinite-long path of the decision process before it reaches $\epsilon$ optimality; see \cite{kakade2003sample}. This differs from our notion of sample complexity under the \SO, which is beyond our current scope. As a result, we do not include these results for comparison in Table 1.

\begin{table}[htbp]
{\footnotesize
\begin{tabular}{|c | p{2.5cm}|c|c|c|c|c}
    \hline
   Method & Setting & Sample Complexity &  Run-Time Complexity & Space Complexity & Reference\\
    \hline
   Phased Q-Learning & $\gamma$ discount factor, $\epsilon$-optimal value & $\frac{\carS \carA }{(1-\gamma)^3\epsilon^2} \ln \frac{1}{\delta}$ &$\frac{\carS \carA }{(1-\gamma)^3\epsilon^2} \ln \frac{1}{\delta}$ & $\carS \carA$& \cite{lattimore2012pac}\\
    \hline
   Model-Based Q-Learning & $\gamma$ discount factor, $\epsilon$-optimal value & $\frac{\carS \carA }{(1-\gamma)^3\epsilon^2} \ln \frac{\carS\carA}{\delta}$ & NA &  $\carS ^2 \carA$ & \cite{azar2012sample}\\ 
	\hline
      Randomized P-D&   $\gamma$ discount factor, $\epsilon$-optimal policy  & $\frac{\carS^3 \carA}{(1-\gamma)^6\epsilon^2}$ & $\frac{\carS^3 \carA}{(1-\gamma)^6\epsilon^2}$
      & $\carS \carA$  & \cite{wang2017randomized} \\
      	\hline
      Randomized P-D &   $\gamma$ discount factor, $\tau$-stationary, $\epsilon$-optimal policy  & $\tau^4 \frac{\S \A { }}{(1-\gamma)^4\epsilon^2} $ & $\tau^4 \frac{\S \A { }}{(1-\gamma)^4\epsilon^2} $ & $\carS \carA$  & \cite{wang2017randomized} \\
        \hline
	Randomized VI & $\gamma$ discount factor, $\epsilon$-optimal policy & $ \frac{|S| |A| \cdot { }}{(1-\gamma)^4 \epsilon^2}$ & $ \frac{|S| |A| \cdot { }}{(1-\gamma)^4 \epsilon^2}$ & $\S\A$  & \cite{soda2017}
	\\       \hline	
	Primal-Dual $\pi$ Learning &   $\tau$-stationary, $\tmix$-mixing, $\epsilon$-optimal policy  & $\frac{(\tau\cdot t^*_{mix})^2 \carS \carA}{\epsilon^2}$ & $\frac{(\tau\cdot t^*_{mix})^2\carS \carA}{\epsilon^2}$ & $\carS \carA$  & This Paper \\
        \hline
\end{tabular}
}\caption{Complexity Results for Sampling-Based Methods for MDP. The sample complexity is measured by the number of queries to the \SO. The run-time complexity is measured by the total run-time complexity under the assumption that each query takes $\tO(1)$ time. The space complexity is the additional space needed by the algorithm in addition to the input.}
\end{table}

Our proposed algorithm and analysis was partly motivated by the stochastic mirror-prox methods for solving convex-concave saddle point problems \cite{nemirovski2005efficient} and variational inequalities \cite{juditsky2011solving}. The idea of stochastic primal-dual update has been used to solve a specific class of minimax bilinear programs in sublinear run time \cite{clarkson2012sublinear}. 
For the discounted MDP, the work \cite{CDC2016} proposed a basic stochastic primal-dual iteration without explicit complexity analysis and later \cite{chen2016stochastic} established a sample complexity upper bound $\cO(\frac{\S^{4.5}\A}{\epsilon^{2}}) $. 
A most relevant prior work is the author's recent paper \cite{wang2017randomized}, which focused on the discounted MDP. The work \cite{wang2017randomized} proposed a randomized mirror-prox method using adaptive transition sampling, which applies to a special saddle point formulation of the Bellman equation. For discounted MDP, it achieved a total runtime/sample complexity of $\tO ( \frac{\S^3 \A { }}{(1-\gamma)^6\epsilon^2} )$ for finding a policy $\pi$ such that $\|v^{\pi} - v^*\|_{\infty}\leq \epsilon$. For discounted MDP such that the stationary distribution satisfies $\tau$-stationarity (see Assumption 1 in the current paper), it finds an approximate policy achieving $\epsilon$ reward  from a particular initial distribution with sample size/run time $\tO (\tau^2\frac{\S \A}{(1-\gamma)^4\epsilon^2}) $.

In this work, we develop the $\pi$ learning method for the case of undiscounted average-reward MDP. Our approach follows from that of \cite{wang2017randomized}, however, our analysis is much more streamlined and applies to the more general undiscounted problems. Without assuming any discount factor, we are able to characterize the complexity upperbound for infinite-horizon MDP using its mixing and stationary properties. Comparing to \cite{wang2017randomized}, the complexity results achieved in the current paper are much sharper, mainly due to the natural simplicity of average-reward Markov processes. To the author's best knowledge, our results provide the first sublinear run time for solving infinite-horizon average-reward MDP without any assumption on discount factor or finite horizon. 


\section{Ergodic MDP, Bellman Equation, and Duality}

Consider an AMDP that is described by a tuple $\mathcal{M}=(\cS,\cA,\cP, \br )$. In this paper, we focus on AMDP that is ergodic (aperiodic and recurrent) under any stationary policy. For a stationary policy $\pi$, we denote by $\nu^{\bmu}$ the stationary distribution of the Markov decision process which satisfies $\lf(P^{\pi}\ri)^{\top} \nu^{\bmu} = \nu^{\bmu}.$ We make the following assumptions on the stationary distributions and mixing times: 
\begin{assumption}[Ergodic Decision Process]\label{assumption-tau}
The Markov decision process specified by $\mathcal{M}=(\cS,\cA,\cP, \br )$ is $\tau$-stationary in the sense that it is ergodic under any stationary policy $\pi$ and there exists $\tau >1$
such that  
$$ \frac1{\sqrt{\tau}\S} \be\leq \nu^{\bmu} \leq \frac{\sqrt{\tau}}{\S} \be.$$
\end{assumption}

Assumption \ref{assumption-tau} characterizes a form of complexity of MDP in terms of the range of its stationary distributions. 
The factor $\tau$ characterizes a notion of complexity of ergodic MDP, i.e., the variation of stationary distributions associated with different policies. Suppose that some policies induce transient states (so the stationary distribution is not bounded away from zero). In this case, we as long as there is some policy that leads to an ergodic process, we can restrict our attention to mixture policies in order to guarantee ergodicity.
In this way, we can always guarantee that Assumption 1 holds on the restricted problem at a cost of some additional approximation error.

\begin{assumption}[Fast Mixing Markov Chains] \label{assumption-mix}
The Markov decision process specified by $\mathcal{M}=(\cS,\cA,\cP, \br )$ is $\tmix$-mixing in the sense that 
$$
t^*_{mix } \geq \max_{\pi} \min\lf\{t\geq1 \mid  \|(P^{\pi})^{t}(i,\cdot) - \nu^{\pi} \|_{TV} \leq \frac14,~\forall i \in\cS \ri\}, 
$$
where $\| \cdot\|_{TV}$ is the total variation.
\end{assumption}

Assumption \ref{assumption-mix} requires that the Markov chains be sufficiently ``rapidly mixing." 
The factor $\tmix$ characterizes how fast the Markov decision process reaches its stationary distribution from any state under any policy. Our results suggest that the $\pi$ learning method would work extremely well on ``rapidly mixing" decision processes where $\tmix$ is a small constant. A typical example is autonomous driving, where the previous actions get forgotten quickly.
On the other hand, the current format of $\pi$ learning might work poorly for problems such as the maze in which the mixing time can be very large and most policies are non-ergodic. This is to be improved.


\vspace{6pt}

Consider an MDP tuple $\mathcal{M}=(\cS,\cA,\cP, \br )$ that satisfies Assumptions 1 and 2.
For a fixed policy $\bmu$, the average reward $\bar v^{\pi} >0 $ is defined as
$$\bar v^{\pi}\equiv \bar v^{\pi} (i)  = \lim_{N\to\infty} \E^{\bmu}\[  \frac1N \sum^{N}_{t=1}     r_{i_{t} i_{t+1} }(a_{t}) ~\Big\vert~ i_1= i\],~i\in\cS,
$$
where $\E^{\bmu}\[ \cdot\]$ is taken over the random state-action trajectory $\{i_1,a_1,i_2,a_2,\ldots\}$ generated by the Markov decision process under policy $\pi$. Note that the average reward $\bar v^{\pi}\equiv \bar v^{\pi} (i)$ is state-invariant, so we treat it as a scalar.

\paragraph{Bellman Equation}
According to the theory of dynamic programming  \cite{puterman2014markov, bertsekas1995dynamic}, the value $\bar v^*$ is the optimal average reward to the AMDP $\cM$ if and only if it satisfies the following $|\cS|\times (|\cS|+1)$ system of equations, known as the {\it Bellman equation}, given by
\begin{equation*}\label{eq-bell}
\begin{split}
\bar v^*+ h^*_i  
= \max_{a\in\cA }&\Bigg\{    \sum_{j\in\cS} p_{ij}(a)h^*_j +  \sum_{j\in\cS}  p_{ij}(a)r_{ij}(a)\Bigg\},\qquad \forall~ i \in \cS,\\
\end{split} 
\end{equation*}
for some vector $\bh^*\in\Re^{\S}$.
A stationary policy $\bmu^*$ is an optimal policy of the AMDP if it attains the elementwise maximization in the Bellman equation (Theorem 8.4.5 \cite{puterman2014markov}). For finite-state AMDP, there always exists at least one optimal policy $\pi^*$. If the optimal policy is unique, it is also a deterministic policy. If there are multiple optimal policies, there exist infinitely many optimal randomized policies.

Note that the preceding Bellman equation has one unique optimal solution $\bar v^*$ but infinitely many solutions $\bh^*$. In the remainder of this paper, we augment the Bellman equation with an additional linear equality constraint 
$$(\nu^{\pi^*})^{\top}\bh^* = 0,$$ 
where $\nu^{\pi^*}$ is the stationary distribution under policy $\pi^*$. Now the augmented Bellman equation has an unique optimal solution. We refer to such a unique solution as the {\it difference-of-value} vector, and we denote it by $\bh^*$ throughout the rest of the paper. The {difference-of-value} vector $\bh^*$ can be informally defined as
$$
 h^{*}_{i} = \lim_{N\to\infty} \E^{\bmu^*}\[  \sum^{N}_{t=1}     r_{i_{t} i_{t+1} }(a_{t}) - N \bar v^{*} ~\Big\vert~ i_1= i\],\qquad \forall~i\in\cS.
$$
It characterizes the transient effect of each initial state under the optimal policy. 

\paragraph{Linear Duality Of The Bellman Equation}
The nonlinear Bellman equation is equivalent to the following $(|\cS|+1)\times (|\cS| |\cA|)$ linear programming problem (see \cite{puterman2014markov} Section 8.8):
\begin{equation}\label{eq-primal}\begin{split}
&\hbox{minimize}_{\bar v,\bh}\ \  \bar v\\
&\hbox{subject to}\ \  \bar v\cdot \be + \left({I} -  P_a\right)  \mathbf{h} - \mathbf{r}_a  \geq 0 , \qquad \forall~a\in \cA,
\end{split}\end{equation}
where $P_a\in\Re^{|\cS|\times |\cS|}$ is the matrix whose $(i,j)$-th entry equals to $p_{ij}(a)$, ${I}$ is the identity matrix with dimension $|\cS|\times |\cS|$, and $\br_a\in\Re^{|\cS|}$ is the expected state-transition reward vector under action $a$, i.e., 
$$r_a(i) = \sum_{j\in\cS} p_{ij}(a) r_{ij}(a),\qquad \forall~ i\in\cS.$$
We associate each constraint of the primal program \eqref{eq-primal} with a dual variable $\mu_a\in\Re^{|\cS|}, a\in\cA$. 
The dual linear program of \eqref{eq-primal} is 
\begin{equation}\label{eq-dual}\begin{split}
&\hbox{maximize} \  \sum_{a\in\cA} \mu_a^{\top} \mathbf{r_a}\\
&\hbox{subject to }  \sum_{a\in\cA} \left({I} - P_a^{\top}\right)  \mu_a = 0, \\
& \qquad\qquad\ \ \suma\sumi\mu_{a,i}=1, \quad \mu_a\geq 0,~~\forall~a\in\cA. 
\end{split}\end{equation}
It is well known that each deterministic policy of the AMDP corresponds to a basic feasible solution to the dual linear program \eqref{eq-dual}. A randomized policy is a mixture of deterministic policies, so it corresponds to a feasible solution of program \eqref{eq-dual}. We denote by $\mu^* = (\mu_a^*)_{a\in\cA}\in\Re^{\S\A}$ the optimal solution to the dual linear program \eqref{eq-dual}. If there is a unique optimal dual solution, it must be a basic feasible solution. In this case, the basis of $\mu^*$ corresponds to an optimal deterministic policy.


\section{Primal-Dual $\pi$ Learning}

In this section, we develop the Primal-Dual $\pi$ Learning Method ({\it $\pi$ learning} for short). Our first step is to examine the nonlinear Bellman equation and formulate it into a bilinear saddle point problem with specially chosen primal and dual constraints. Our second step is to develop the Primal-Dual $\pi$ Learning method and discuss its implementation and run-time complexity per iteration.

\subsection{Saddle Point Formulation of Bellman Equation}

In light of linear duality, we formulate the linear programs \eqref{eq-primal}-\eqref{eq-dual} into an equivalent minimax problem, given by
\begin{equation*}
\begin{split}
&\min_{\bar v, \bh} \max_{\mu\geq 0}  \bar v + \suma \bpi_a^{\top} \lf( - \bar v\cdot \be + (  P_a - I ) \bh + \br_a\ri).
\end{split}
\end{equation*}
The minimax formulation is more preferable to the linear program formulation because it has much simpler constraints. 
We construct the sets $\cH$ and $ \cU$ to be the {\it search spaces} for the value and the policy, respectively, given by
$$
\mathcal{H} 
= 
\lf\{ \bh\in\Re^{\S} ~\Big\vert~ \|\bh\|_{\infty}\leq 2 t^*_{mix} \ri\},$$
and
\begin{align*}
 \cU = \lf\{ \mu = (\mu_a)_{a\in\cA}~\Big\vert~ \be^{\top}\mu =1, \mu\geq 0, \suma \mu_a \geq \frac1{\sqrt{\tau}\S}\be \ri\}.
\end{align*}
Since $ \suma \bpi_a^{\top}\be = \bpi^{\top}\be= 1$, we simplify the minimax problem to
\begin{equation}\label{eq-saddle}
\begin{split}
&\min_{\bh \in\cH} \max_{\mu \in\cU}  \suma \bpi_a^{\top} \lf(  (  P_a - I ) \bh + \br_a\ri).
\end{split}
\end{equation}
The search space for the dual vector given by $\cU$ essentially reflects Assumption 1. Recall that Assumption 1 suggests that the stationary distribution of any policy belongs to a certain range, therefore it is sufficient to search for the dual variable within that range.
The search space for the difference-of-value vector given by $\mathcal{H}$ essentially reflects Assumption 2 on the fast mixing property of the MDP. The fast mixing condition implies that one can move from any state to any state within a bounded number of steps, therefore the relative difference in their values is bounded by the expected traverse time.  In what follows, we verify that $\bh^*\in \cH$ and $\mu^*\in\cU$ under Assumptions 1 and 2.
\begin{lemma}\label{lemma-HU}
Under Assumptions 1 and 2, the optimal primal and dual solutions $\bar v^*,\bh^*,\mu^*$ to the linear programs \eqref{eq-primal}-\eqref{eq-dual} satisfy $\bar v^*\in[0,1]$, $\bh^*\in \cH$ and $\mu^*\in\cU$.
\end{lemma}
\pf
Since $\bar v^*$ is the average reward under $\pi^*$ and each reward per period belongs to $[0,1]$, we obtain that $\bar v^*\in[0,1].$

Let $P^*$ be the transition probability matrix under $\pi^*$.
Let $\nu^*$ be the stationary distribution under $\pi^*$, so the difference-of-value vector $\bh^*$ satisfies $(\bh^*)^{\top}\nu^* = 0$.
Let $\Pi$ be the matrix with all rows equaling to $(\nu^*)^{\top}$, therefore $\Pi \bh^* = 0 $.
Letting $m=\tmix$, we have $\|(P^*)^m(i,\cdot) - \pi^*\|_{TV}\leq 1/4$ for all $i\in\cS$, therefore $ \|(P^*)^m - \Pi\|_{\infty} \leq 1/4$. We apply the relation $\bh^* = P^* \bh^* + \br^* -\bar v\cdot\be $ inductively for $m$ times, use $\Pi \bh^* = 0 $ and obtain
$$\bh^* = \sum^{m-1}_{k=0} (P^*)^k \br + (P^*)^m \bh^* - m \bar v^*\cdot\be 
=\sum^{m-1}_{k=0} \lf((P^*)^k \br - \bar v^*\cdot\be  \ri) +\lf( (P^*)^m  -\Pi\ri)\bh^*  .$$
We take $\|\cdot\|_{\infty}$ on both sides of the above, use the triangle inequality and obtain
$$\|\bh^*\|_{\infty} \leq \sum^{m-1}_{k=0} \| (P^*)^k \br -  \bar v^* \|_{\infty} + \|(P^*)^m - \Pi\|_{\infty} \|\bh^*\|_{\infty}  
\leq m + \frac14 \|\bh^*\|_{\infty}  $$
It follows that $\|\bh^*\|_{\infty} \leq (4/3)m \leq 2\tmix$ and $\bh^*\in\cH$.

Recall $\mu^*$ is the optimal dual solution to the linear programs \eqref{eq-primal}-\eqref{eq-dual}. The dual feasibility of $\mu^*$ suggests that
$0 = \sum_{a\in\cA} \left({I} - P_a^{\top}\right)  \mu_a^* =  \left({I} - (P^*){\top}\right) \sum_{a\in\cA} \mu_a^* $, therefore $ \sum_{a\in\cA} \mu_a^*$ is the stationary distribution corresponding to the transition matrix $P^*$ under the optimal policy $\pi^*$. It follows from Assumption \ref{assumption-tau} that $\mu^*\in\cU$.
\qed

 \subsection{The Primal Dual $\pi$ Learning Algorithm}

\def\bd{\mathbf{d}}

Motivated by the minimax formulation of the Bellman equation, we propose the Primal-Dual $\pi$ Learning method as follows:
The $\pi$ learning method makes iterative updates to a sequence of primal and dual variables $\{\mu^t,\bh^t\}_{t=0}^T$.
At the $(t+1)$ iteration, the algorithm draws a random state-action pair $(i,a)$ with probability $\mu^t_{i,a}$ and query the \SO\ for a state transition to a random next state $j$ with probability $p_{ij}(a)$. Then the $\pi$ learning method updates according to
\begin{equation}\label{eq-gg}
\begin{split}
\mu^{t+1} &= \argmin_{\mu\in \cU_{.}}\KL(\mu ||\mu^{t} \cdot \exp(\g^{t+1})),
\\
\bh^{t+1}& = \mathbf{Proj}_{\cH} \[\bh^t +  \bd^{t+1} \] ,
\end{split}
\end{equation}
where ``$\cdot$" denotes elementwise multiplication, $\mathbf{Proj}_{\cH}$ denotes the Euclidean projection onto $\cH$, $\Delta^{t+1}\in\Re^{\S\A},$ $\bd^{t+1}\in\Re^{\S}$ are random vectors generated conditioned on $\mu^t,\bh^t$ according to
\begin{equation}\label{eq-gg}
\begin{split}
\Delta^{t+1} \mid  \cF_t 
&= {\beta}  \cdot \frac{h^t_j - h^t_{i} +r_{ij}(a) -M }{  \mu^t_{{i,a}}} \mathbf{e}_{i,a},\qquad 
\hbox{with probability } \mu^t_{i,a},
\\
\bd^{t+1} \mid  \cF_t  &= \alpha\cdot ( \mathbf{e}_{i} - \mathbf{e}_j),\qquad 
\hbox{with probability }
 \mu^t_{i,a}p_{i,j}(a),
\end{split}
\end{equation}
where we use $\cF_t$ to denote the collection of all random variables up to the $t$-th iteration.
We note that $\Delta^{t+1} $ is a vector of dimension $\S\A$ but it only has one single nonzero entry. Similarly, $\bd^{t+1}$ is a vector of dimension $\S$ but it has only two nonzero entries, whose coordinates are randomly generated by sampling a single state transition.
We can easily verify that
$$\E\[\Delta^{t+1}_a \mid  \cF_t \] = \beta \lf((P_a-I)\bh^{t} +\br_a - M \cdot \be\ri),\qquad a\in\cA,$$
and 
$$
\E\[\bd^{t+1} \mid  \cF_t \] = \alpha \suma \mu_a^{\top} (I-P_a).
$$
In other words, the primal and dual updates $\Delta^{t+1},\bd^{t+1}$ are conditionally unbiased partial derivatives of the minimax objective. 

\subsection{Implementations and Fast Updates}

Let us consider how to implement the $\pi$ learning method in order to minimize the run time per iteration. We define the auxilary variables ${\xi^t} = \lf({\xi}^t_{i}\ri)_{i\in\cS}$, $\pi= \lf(\pi^t_{i,a}\ri)_{i\in\cS,a\in\cA}$ such that 
$$\xi_i^t = \suma \mu_{i,a}^t,\qquad \pi^t_{i,a}= \frac{\mu^t_{i,a}}{\xi_i^t},\qquad
{\mu}^t_{i,a} = \xi^t_i \pi^t_{i,a}\qquad
 \forall~i\in\cS,a\in\cA.$$
 Note that
$\xi^t$ is a vector of probability over states, and $\pi^t$ is a randomized stationary policy that specifies the probability distribution for choosing actions at each given state. We implement the Primal-Dual $\pi$ Learning method given by iteration \eqref{eq-gg} in Algorithm 1. 

Now we analyze the computational complexity of Algorithm 1. Each iteration draws one state-action-state triplet from the \SO. The updates on $\bh$ are made to two coordinates, thus taking $\tO(1)$ time. The updates on $\pi$ are multiplicative, which take $\tO(1)$ time if $\pi$ is represented using convenient data structures like binary trees (see Prop.\ 1 of \cite{wang2017randomized}). The updates on $\xi$ involve information projection onto the set $\{\xi \geq \frac1{\sqrt{\tau}\S}\be, \xi^{\top} \be = 1,\xi \geq 0\}$.
This can be done by maintaining and updating the shifted vector $ \xi - \frac1{\sqrt{\tau}\S}\be$ using a binary-tree structure.  This idea is also used in the algorithm implementation of \cite{wang2017randomized}. Accordingly, Step 10 of Algorithm 1 takes $\tO(1)$ run time. To sum up, each iteration of Algorithm 1 draws one sample transition and makes updates in $\tO(1)$ time. 
The space complexity of Algorithm 1 is $\cO(\S\A)$ space, mainly to keep track of $\pi$ and its running average.


\begin{algorithm}[h!]
\caption{Primal-Dual $\pi$ Learning}\label{algo:primaldual}
\begin{algorithmic}[1]
\State {\bf Input:} Precision level $\epsilon>0$, $\cS$, $\cA$, $\tmix,\tau$, \SO
\State Set $\bh = 0\in\Re^{\S}$, $\xi = \frac1{\S} \be \in\Re^{\S} $, $\pi_i =\frac1{\A}\be\in\Re^{\A}$ for all $i\in \cS$
\State Set $T = \tau^2 (\tmix)^2 \S\A$
\State  Set $\beta= \frac1{\tmix}\sqrt{\frac{\log \lf(\S\A \ri)}{2\S\A T}}, \alpha= \S \tmix \sqrt{\frac{\log \lf(\S\A\ri)}{2\A T}} ,  M = 4 \tmix+1$
\For{$t = 1,2,3,\ldots,T $}
\State Sample $(i,a)$ with probability $\xi_i \pi_{i,a}$
\State Sample $j$ with probability $p_{ij}(a)$ from \SO
\State
$\g \leftarrow   {\beta}  \cdot \frac{ h^t_j - h^t_i  +r_{ij}(a) - M }{ \xi^t_{i,a}\pi^t_{i,a}} $
\State
$h_{i}  \leftarrow \min\{h_{i} +  
{\alpha} ,2\tmix \} ,
h_j \leftarrow \max\{
h_j -  {\alpha}  ,- 2\tmix \}  , $
\State
$\xi_i \leftarrow \xi_i +\pi_{i,a} \lf(\exp\lf\{ \g \ri\}-1\ri),
\xi \leftarrow  \hbox{argmin}_{\hat\xi} \lf\{\KL(\hat \xi || \xi) \mid \be^{\top}\hat\xi = 1,\hat\xi\geq0, \hat\xi\geq \frac1{\sqrt{\tau}\S} \be \ri\}$
\State
$\bmu_{i,a} \leftarrow \bmu_{i,a} \cdot \exp\lf\{\g \ri\},
\bmu_i \leftarrow \bmu_i/\|\bmu_i\|_1$

\State  $ \pi^{t+1} \leftarrow \pi$
\State $t \leftarrow t+1$
\EndFor

\State\label{step-output}{\bf Ouput:} $\hat \bmu=  {\frac1T\sum^{T}_{t=1} \bmu^t }$
\end{algorithmic}
\end{algorithm}



\def\bw{\mathbf{w}}

\section{Sample Complexity and Run Time Analysis}

In this section, we establish the sample complexity for the Primal-Dual $\pi$ Learning method given by Algorithms 1. We also show that Algorithm 1 applies to the computation problem of MDP and gives a sublinear run-time algorithm. 


\subsection{Primal-Dual Convergence}

Each iteration of Algorithm 1 performs a primal-dual update for the minimax problem \eqref{eq-saddle}. Our first result concerns the convergence of the primal-dual iteration.

\begin{theorem}[\bf Finite-Iteration Duality Gap] \label{thm-dualitygap}
Let $\mathcal{M}=(\cS,\cA,\cP, \br )$ be an arbitrary MDP tuple satisfying Assumptions 1, 2.  
Then the sequence of iterates generated by Algorithm 1 satisfies
$$ \frac1T\sum^T_{t=1}\E\[\suma ( \bh^* - \a P_{a} \bh^* - \br_{a} )^{\top}\mu^t_{{a}}  \] + \bar v^* \leq \tO\lf( \tmix \sqrt{\frac{ \S\A }{T}}\ri) ,
$$
where $\mu^t_{i,a} =\xi^t_{i}\pi^t_{i,a} $ for $i\in\cS,a\in\cA$, $t=1,\ldots,T$.
\end{theorem}

Theorem \ref{thm-dualitygap} establishes a finite-time error bound of a particular ``duality gap." It characterizes the level of violation of the linear complementarity condition.  Our proof shares a similar spirit as that of Theorem 1 in \cite{wang2017randomized}. Note that the analysis of \cite{wang2017randomized} does not easily extend to the average-reward MDP and the $\pi$ learning method. As a result, we have to develop a separate new convergence analysis. The complete proof is established through a series of lemmas, which we defer to Appendix.


\subsection{Sample Complexity for Achieving $\epsilon$-Optimal Policies}

We have shown that the expected duality gap diminishes at a certain rate as Algorithm 1 iterates. It remains to analyze how many time steps are needed for the duality gap to become sufficiently small, and how a small duality gap would imply a near-optimal policy. We obtain the following result. 

\begin{lemma} \label{lemma-avgv}
For any policy $\pi$, its stationary distribution $\nu^{\pi}$ and average reward $\bar v^{\pi}$ satisfies
\begin{align*}
\bar v^{\pi}
={\lf(\nu^{ \pi}\ri)}^{\top} \suma \mathbf{diag}(\pi_a) ((P_a - I)\bh^*+\br_a),
\end{align*}
and 
$$
\bar v^* - \bar v^{\pi}
={\lf(\nu^{ \pi}\ri)}^{\top} \suma \mathbf{diag}(\pi_a) (\bar v^* \cdot \be+ (I - P_a )\bh^*- \br_a).
$$
\end{lemma}
\pf
Consider an arbitrary policy $\pi$. Let $\nu^{ \pi}$ be the stationary distribution under policy $\pi$, so we have $\lf(\nu^{ \pi}\ri)^{\top} P^{\pi}  = \lf(\nu^{ \pi}\ri)^{\top}$. Then we obtain the first result
\begin{align*}
\bar v^{\pi}
= {\lf(\nu^{ \pi}\ri)}^{\top} \br^{\pi}
= {\lf(\nu^{ \pi}\ri)}^{\top}  ((P^{\pi} - I)\bh^*+\br^{\pi} )
={\lf(\nu^{ \pi}\ri)}^{\top} \suma \mathbf{diag}(\pi_a) ((P_a - I)\bh^*+\br_a).
\end{align*}
Using the fact that
${\lf(\nu^{ \pi}\ri)}^{\top} \suma \mathbf{diag}(\pi_a)  \be = 1$, we obtain the second result.
\qed

Now we are ready to show that the $\pi$ learning method outputs an approximate policy whose average reward is close to the optimal average reward. Our second main result is as follows.

\begin{theorem}[\bf Sample Complexity of Single-Run $\pi$ Learning (Algorithm 1)] \label{thm-complexity1}
Let $\mathcal{M}=(\cS,\cA,\cP, \br )$ be an arbitrary MDP tuple satisfying Assumptions 1, 2, let $\epsilon>0$. Then by letting Algorithm 1 run for the following number of iterations/samples
$$ T  =  \Omega \lf((\tau\cdot t^*_{mix})^2\cdot \frac{ \S \A }{\epsilon^2}\ri)$$
it outputs an approximate policy $\hat \pi$ such that
$ \bar v^{\hat\pi} \geq \bar v^* - \epsilon$ with probability at least $2/3$.
\end{theorem}

\pf
Consider the policy given by $\hat \pi = \frac1T\sum^T_{t=1} \pi^t.$ 
Note that $\frac{1}{\sqrt{\tau}\S}\be \leq \nu^{\hat\pi} \leq \frac{\sqrt{\tau}}{\S}\be$ (by Assumption \ref{assumption-tau}) and $\frac{1}{\sqrt{\tau}\S}\be \leq \xi^t $ (since $\mu^t\in\cH$). Then we have
$$\nu^{\hat\pi} \leq \frac{\sqrt{\tau}}{\S}\be = \tau\cdot \frac1{\sqrt{\tau} \S}\be \leq \tau \xi^t.$$ 
According to Lemma \ref{lemma-avgv}, we have
\begin{align*}
\bar v^*   - \bar v^{\hat \pi} 
&=  {\lf(\nu^{ \hat \pi}\ri)}^{\top} \suma \mathbf{diag}(\hat\pi_a)  (\bar v^*\cdot \be  + (I-P_a )\bh^* -\br_a )
\\
&= \frac1T\sum^T_{t=1} {\lf(\nu^{\hat \pi}\ri)}^{\top} \suma  \mathbf{diag}(\pi^t_a)   (\bar v^*\cdot \be  + (I-P_a )\bh^* -\br_a )
\\
&\leq \tau\cdot \frac1T\sum^T_{t=1} {\lf(\xi^t\ri)}^{\top} \suma \mathbf{diag}(\pi^t_a) (\bar v^*\cdot \be  + (I-P_a )\bh^* -\br_a )
\\
&= 
\tau \cdot\lf(\frac{1}T\sum^T_{t=1}\sumia \mu^t_{i,a}( \bh^* - \a P_{a} \bh^* - \br_{a} )_i  + \bar v\ri)
,
\end{align*}
where the inequality uses the fact $\nu^{\hat\pi} \leq  \tau \xi^t$ for all $t$ (due to the dual constraint $\cH$) and the primal feasibility $(\bar v^*\cdot \be  + (I-P_a )\bh^* -\br_a )
\geq 0$ for all $a\in\cA$.
We use the Markov inequality and obtain that 
$$\bar v^*  - \bar v^{\hat \pi} \leq \frac{3\tau}{2}
\lf(\frac1T\sum^T_{t=1}\E\[\suma ( \bh^* - \a P_{a} \bh^* - \br_{a} )^{\top} \mu^t_{a}  \] + \bar v^*\ri) $$with probability at least $2/3$. 
Now if we pick $T = \Omega(\tau^2 (\tmix)^2 \frac{\S \A}{\epsilon^2})$ and apply Theorem \ref{thm-dualitygap}, we obtain that $\bar v^{\hat \pi} \geq \bar v^* -  \epsilon$ with probability at least $2/3.$
\qed

\subsection{Boosting The Success Probability to $1-\delta$}

Our next aim is to achieve an $\epsilon$-optimal policy with probability that is arbitrarily close to 1. To do this, we need to run Algorithm 1 for sufficiently many trials and pick the best outcome. This requires us be able to evaluate multiple candidate policies and select the best one out of many. 
In the next lemma, we show that it is possible to approximately evaluate any policy $\pi$ within $\epsilon$-precision using $\tO(\frac{\tmix}{\epsilon^2})$ samples.

\begin{lemma}[Approximate Policy Evaluation]\label{lemma-pe}
There exists an algorithm that outputs an approximate value $\bar Y$ such that $\bar \bv^{\pi}  -  {\epsilon} \leq \bar Y \leq  \bar \bv^{\pi}+\epsilon $ with probability at least $1-\delta$ in 
$\tO(\frac{\tmix}{\epsilon^2}\log(\frac1{\delta})) $ time steps.
\end{lemma}
\pf 
Consider the algorithm that generates a sequence of $L$ consectutive state transitions according to the \SO and outputs the empirical mean reward, which we denote by $\bar Y$.
Note that $\bar Y$ is the empirical mean of $L$ Markov random variables in $[0,1]$.
We apply the McDiarmid inequality for Markov chains to the $L$-step empirical reward $\bar Y$ and obtain
$$\mathbf{P}( |\bar Y - \bar v^{\pi}| \geq \epsilon) \leq 2\exp\lf( -\frac{ L\epsilon^2}{\tmix}\ri) $$
When $L \geq \frac{\tmix}{\epsilon^2} \log(\frac1\delta)$, we have $ |\bar Y - \bar v^{\pi}| <\epsilon$ with probability at least $1-\delta.$
\qed

\vspace{6pt}


Now we prove that by repeatedly running Algorithm 1 and using approximate policy evaluation, one can compute a near-optimal policy with probability arbitrarily close to 1.  The main arguments are (1) the best policy out of multiple trials must be close-to-optimal with high probability; (2) the policy evaluation is nearly accurate with high probability, therefore the output policy (which performs the best in policy evaluation) is also close-to-optimal.
Our main result is as follows. 

\begin{theorem}[\bf Overall Sample Complexity]\label{theo:rates2}
Let $\mathcal{M}=(\cS,\cA,\cP, \br )$ be an arbitrary MDP tuple satisfying Assumptions 1, 2 and let $\epsilon>0$ and $\delta\in(0,1)$ be arbitrary values.  
Then there exists an algorithm that draws the following number of state transitions
$$ T  =  \Omega \lf((\tau\cdot t^*_{mix})^2 \cdot \frac{\S \A }{\epsilon^2} \log\frac1{\delta} \ri)$$
and outputs an approximate policy $\hat \pi$ such that
$ \bar v^{\hat\pi} \geq \bar v^* - \epsilon$ with probability at least $1-\delta$. 
\end{theorem}


\pf
We describe an approach that runs Algorithm 1 for multiple times in order to achieve an $\epsilon$-optimal policy with probability $1-\delta$:
\begin{enumerate}
\item
We first run Algorithm 1 for $K$ independent trials with precision parameter $\frac{\epsilon}3$, and we denote the output policies by 
$\bmu^{(1)},\ldots, \bmu^{(K)}$. The total running time is $K\cdot N_{\frac{\epsilon}3}$, where $N_{\frac{\epsilon}3}$ is the number of samples needed by Algorithm 1. According to Theorem 1, each trial generates an $\epsilon/3$-optimal policy with probability at least $2/3$. 

\item For each output policy $\pi^{(k)}$, we conduct approximate value evaluation for $L$ time steps and obtain an approximate evaluation $\bar Y^{(k)}$ with precision level $\frac{\epsilon}{3}$ and fail probability $\frac{\delta}{2K}$. According to Lemma \ref{lemma-pe}, we have
$$  \bar Y^{(k)} - \bar \bv^{\pi^{(k)}}\in  [- \frac{\epsilon}3  ,  \frac{\epsilon}3] ,
$$ 
with probability at least $1-\frac{\delta}{2 K}$, and this step takes $K\cdot L = K\cdot  \tO(\frac{\tmix}{\epsilon^2} \log\lf(\frac{K}\delta\ri)) $ time steps.

\item Output $\hat\pi = \bmu^{(k^*)}$ such that $k^* = \hbox{argmax}_{k=1,\ldots,K} \bar Y^{(k)}$. 
\end{enumerate}
The number of samples required by the above procedure is $\tO(N_{\frac{\epsilon}3} \log\frac1\delta + L \log\frac1\delta )$. The space complexity is $\cO(\S\A).$

Now we verify that $\hat\pi$ is indeed near-optimal with probability at least $1-\delta$, as long as $K$ is chosen appropriately.
Let $\mathcal{K} =\lf\{k\in[K]\mid \bar\bv^{\pi^{(k)}} \geq  \bar\bv^{*} - \frac{\epsilon}3\ri\} $, which can be interpreted as the set of successful trails of Algorithm 1. 
Consider the event where $\mathcal{K}\neq\emptyset$ and all policy evaluation errors belong to the small interval $[-\frac{\epsilon}3 , \frac{\epsilon}3]$. In this case, we have $ \bar \bv^{\pi^{(k)}} -\frac{\epsilon}3\leq \bar Y^{(k)} \leq \bar \bv^{\pi^{(k)}} +\frac{\epsilon}3$ for all $k$ and $\bv^{\pi^{(k)}}  \geq  \bar\bv^{*} -\frac13\epsilon$ if $k\in\mathcal{K}$. As long as $\mathcal{K}$ is nonempty, the output policy which has the largest value of $\bar Y^{(k)} $ must satisfy  $\bar Y^{(k)} \geq  \bar\bv^{*} -\frac23\epsilon$. Since the policy evaluation error is bounded by $\frac{\epsilon}3$, it follows that this policy must be $\epsilon$-optimal.
We use the union bound to obtain 
\begin{align*}
\mathbf{P}\lf( \bar\bv^{\hat\pi} <   \bar\bv^{*} - \epsilon \ri)
&\leq  \mathbf{P}\lf(\lf\{ \mathcal{K} =\emptyset \ri\} \cup \lf\{ \exists k: \bar Y^{(k)} - \bar \bv^{\pi^{(k)}}\notin [- \frac{\epsilon}3 ,\frac{\epsilon}3] \ri\}\ri)
\\
&\leq   \mathbf{P} \lf( \mathcal{K} =\emptyset \ri) +
\mathbf{P}\lf(  \exists k: \bar Y^{(k)} - \bar \bv^{\pi^{(k)}}\notin [- \frac{\epsilon}3 ,\frac{\epsilon}3] \ri)
\\
&\leq \prod_{k=1}^K \mathbf{P} \lf(  \bar \bv^{\pi^{(k)}} <  \bar\bv^{*}- \frac{\epsilon}3 \ri)
\\
&~~+
\sum_{k=1}^K \mathbf{P}\lf(\bar Y^{(k)} - \bar \bv^{\pi^{(k)}}\notin [- \frac{\epsilon}3  ,\frac{\epsilon}3] \ri)
\\
&\leq (1/3)^K +  K\cdot \frac{\delta}{2K} .
\end{align*}
By choosing $K  = \log(2/\delta) \geq \log_{1/3}(\frac{\delta}2)$, we obtain $\mathbf{P}\lf( \bar\bv^{\hat\pi} <  \bar\bv^{*} -\epsilon \ri)\leq \delta$. Then the output policy $\hat \pi$ is $\epsilon$-optimal with probability at least $1-\delta$. 
\qed

The $\pi$ learning method is not only useful in the setting of reinforcement learning. It also applies to the computational problem of approximating the optimal policy when the MDP model is explicitly given. We obtain the following sublinear run-time complexity for numerically solving the AMDP.

\begin{theorem}[\bf Sublinear Run Time for Ergodic MDP]\label{thm-runtime}
Let the $\mathcal{M}=(\cS,\cA,\cP, \br )$ be an MDP tuple that is specified in data structures that enable sampling state transitions in $\tO(1)$ time. Then there exists an algorithm that takes $\mathcal{M}$ as the input and outputs an approximate policy $\hat \pi$ such that
$ \bar v^{\hat\pi} \geq \bar v^* - \epsilon$ with probability at least $1-\delta$ in run time
$$\Omega \lf((\tau\cdot t^*_{mix})^2\cdot \frac{\S \A }{\epsilon^2} \log\frac1{\delta} \ri).$$
\end{theorem}
\pf
Note that each iteration of Algorithm 1 draws one sample from the \SO\ and makes updates in $\tO(1)$ time. Then the result of Theorem \ref{thm-runtime} follows straightforwardly from the sample complexity result of Theorem \ref{theo:rates2}. \qed

Theorem \ref{thm-runtime} suggests that one can approximately solve the AMDP problem in sublinear time. In particular, finding an approximately optimal policy does not even require reading most of the input entries.
Remarkably, the sample complexity and run-time complexity of the $\pi$ learning method happen to be equvalent to each other. Such an equivalence holds because the $\pi$ learning method uses each new sample transition in a most computationally efficient way - making only a few coordinate updates to the value and policy vectors. Note that for general reinforcement learning methods, the sample complexity and run-time complexity are typically far from equal to each other.

\section{Summary} 

We have developed a primal-dual $\pi$ learning method for solving the undiscounted ergodic Markov decision problems by sampling state-to-state transitions. The method directly updates the value and policy estimates as new state transitions are observed. This method is model-free and can be implemented efficiently in $\cO(\S\A) $ space. We show that it achieves a sample complexity $\cO((\tau\cdot\tmix)^2\frac{\S\A}{\epsilon^2})$ for ergodic average-reward Markov decision process, where $\tau$ is a parameter characterizing the range of stationary distributions and $\tmix$ is an upper bound of mixing times across all policies. 

The $\pi$ learning method can be applied to approximating the optimal policy when the MDP is fully specified. When state transitions can be sampled in $\cO(1)$ time, the $\pi$ learning method can be used as a randomized algorithm and computes an $\epsilon$-optimal policy in run time $\cO((\tau\cdot \tmix)^2\frac{\S\A}{\epsilon^2})$, which is sublinear with respect to the input size. An open question is to investigate the roles of $\tau,\tmix$ in the complexity and potentially improve the complexity's dependence on these parameters.


 \bibliography{refbib/mdp,refbib/lowerbound,mdp,mdp2,nips_1705}
{\bibliographystyle{plain}}

\appendix

\newpage
\section{Proof of Theorem \ref{thm-dualitygap}: Duality Gap Analysis}\label{sec-lemmas}
 
In this section, we analyze the convergence of Algorithm 1. In what follows, we denote by $\cF_t$ the collection of random variables that are revealed up to the end of the $t$-th iteration. For two probability distributions $p,q$ over a finite set $X$, we denote by $\KL(p||q) = \sum_{x\in X} p(x) \log \frac{p(x)}{q(x)}$ the Kullback-Leibler divergence. We assume that Assumptions 1 and 2 hold throughout the analysis.
We let 
$$\br_a =\sumj p_{ij}(a) r_{ij}(a), ~~ \br^{\pi} = \suma \pi_i(a) \sumj p_{ij}(a) r_{ij}(a), ~~\br^* = \suma \pi^*_i(a) \sumj p_{ij}(a) r_{ij}(a) .$$ We denote by $P^*=P^{\pi^*}$ the transition matrix under the optimal policy. Note that
$$P^{\pi} =\suma \mathbf{diag}({\pi}_a) P_a,\qquad P^{*} =\suma \mathbf{diag}({\pi^*_a}) P_a,$$
where $ \mathbf{diag}({\pi_a}) $ is the diagonal matrix with $\pi_{1,a},\pi_{2,a},\ldots,\pi_{\S,a}$ along its diagonal.


In addition,  
the updates on $\xi^t$ and $\pi^t$ can be equivalently written as updates on ${\mu}^t$, given by
\begin{equation}\label{eq-update}
\begin{split}
{\mu}^{t+1/2}_{{i,a}} &= \frac{{\mu}^{t}_{{i,a}} \cdot \exp(\g^{t+1}_{{i,a}})}{\sum_{i',a'}
{\mu}^{t}_{i',a'} \cdot \exp(\g^{t+1}_{i',a'})},
\qquad \forall\ i\in\cS,a\in\cA,
\\
{\mu}^{t+1}\ \ \ &= \argmin_{\mu\in \cU}\KL(\mu ||\mu^{t+1/2}),
\end{split}
\end{equation}
One can verify that $\mu^t\in\cU$ and $\bh^t\in\cH$ for all $t$ with probability 1.

\begin{lemma} \label{lemma-KL}The iterates generated by Algorithm 1 satisfy
\begin{equation}\label{eq-KL0}
\begin{split}
& \E\[\KL({\mu}^* || {\mu}^{t+1}) \mid \cF_t\] - \KL({\mu}^* || {\mu}^t)  
\leq \sumia ({\mu}^t_{{i,a}}  - {\mu}^*_{{i,a}}) \E\[  \g^{t+1}_{{i,a}} \mid \cF_t\]+  \frac12\sumia {\mu}^t_{{i,a}}\E\[  \lf(\g^{t+1}_{{i,a}}\ri)^2\mid \cF_t\] ,
\end{split}
\end{equation}
for all $t$, with probability 1.
\end{lemma}

\pf
\def\bg{\mathbf{g}}
By using the relation \eqref{eq-update}, we have
\begin{equation}\label{eq-01}\begin{split}
\KL({\mu}^* || {\mu}^{t+1/2}) - \KL({\mu}^* || {\mu}^t) 
&= \sumia {\mu}^*_{{i,a}} \log \frac{{\mu}^*_{{i,a}}}{{\mu}^{t+1/2}_{{i,a}}} - \sumia {\mu}^*_{{i,a}} \log \frac{{\mu}^*_{{i,a}}}{{\mu}^t_{{i,a}}} \\
&= \sumia {\mu}^*_{{i,a}} \log \frac{{\mu}^t_{{i,a}}} {{\mu}^{t+1/2}_{{i,a}}} \\
&= \sumia {\mu}^*_{{i,a}} \log \frac{Z } {\exp( \g^{t+1}_{{i,a}})} \\
&= \sumia {\mu}^*_{{i,a}} \log \lf(Z\ri)  - \sumia {\mu}^*_{{i,a}} \g^{t+1}_{{i,a}}\\
&=  \log Z  -\sumia {\mu}^*_{{i,a}} \g^{t+1}_{{i,a}},
\end{split}\end{equation}
where $Z = \sumia {\mu}^t_{{i,a}} \exp(\g^{t+1}_{{i,a}} )$.
According to \eqref{eq-gg}, we have $  h^t_j - h^t_{i} +r_{ij}(a) -M \leq  2\tmix -(-2\tmix )+1 - 4\tmix - 1\leq 0$ because $h^t_i\in[-2\tmix,2\tmix]$, $r_{ij}(a)\in[0,1]$ and $M=4\tmix+1$. It follows that $ \g^{t+1}_{i,a} \leq 0$ for all $i\in\cS,a\in\cA$ with probability 1. Then we derive
\begin{equation}\label{eq-02}\begin{split}
\log Z =  \log \lf( \sumia {\mu}^t_{{i,a}} \exp(\g^{t+1}_{{i,a}} )\ri) 
&\leq \log \sumia {\mu}^t_{{i,a}} \lf(1+ \g^{t+1}_{{i,a}} + \frac12 \lf(\g^{t+1}_{{i,a}}\ri)^2 \ri)\\
&= \log \lf( 1 + \sumia {\mu}^t_{{i,a}}  \g^{t+1}_{{i,a}} +\frac12 \sumia {\mu}^t_{{i,a}}  \lf(\g^{t+1}_{{i,a}}\ri)^2 \ri)\\
&\leq \sumia {\mu}^t_{{i,a}}  \g^{t+1}_{{i,a}} + \frac12 \sumia {\mu}^t_{{i,a}}  \lf(\g^{t+1}_{{i,a}}\ri)^2 ,
\end{split}\end{equation}
where the first inequality uses the fact $e^x\leq 1+x+\frac12 x^2$ if $x\leq 0$ and the second inequality uses the fact $\log(1+x)\leq x$ for all $x$.
We combine \eqref{eq-01} and \eqref{eq-02} and take conditional expectation $\E\[\cdot\mid \cF_t\]$ on both sides, then we obtain
\begin{equation}
\begin{split}
& \E\[\KL({\mu}^* || {\mu}^{t+1/2}) \mid \cF_t\] - \KL({\mu}^* || {\mu}^t)  \\
& \leq \sumia ({\mu}^t_{{i,a}}  - {\mu}^*_{{i,a}}) \E\[  \g^{t+1}_{{i,a}} \mid \cF_t\]+  \frac12\sumia {\mu}^t_{{i,a}}\E\[  \lf(\g^{t+1}_{{i,a}}\ri)^2\mid \cF_t\] ,
\end{split}
\end{equation}
Finally, we note that $\KL({\mu}^* || {\mu}^{t+1}) \leq \KL({\mu}^* || {\mu}^{t+1/2})$ due to the information projection step (see \cite{cover2012elements} Theorem 11.6.1 on page 367) and that $\mu^*\in\cU$. By combining the preceding two relations, we have obtained \eqref{eq-KL0}.
\qed

\vspace{5pt}

\begin{lemma}\label{lemma-variance}
The iterates generated by Algorithm 1 satisfy 
$$\sumia \bpi^t_{{i,a}}\E\[  \lf(\g^{t+1}_{{i,a}}\ri)^2\mid \cF_t \] \leq  {4 \S \A  (4\tmix+1)^2 \beta^2},$$
for all $t\geq 1$ with probability 1.
\end{lemma}
\pf We have
\begin{align*}
\sumia  \bpi^t_{{i,a}} \E\[  \lf( \g^{t+1}_{{i,a}}\ri)^2\mid \cF_t \]
&= \sumi\suma \bpi^t_{{i,a}} \cdot \xi^t_i  \pi^t_{{i,a}} \cdot  \sumj p_{ij}(a)\lf( {\beta}  \cdot \frac{(  h^t_j - h^t_{i} +r_{ij}(a)   -M )}{ \xi^t_i  \pi^t_{{i,a}}}\ri)^2
\\
&= \sumi \suma \sumj p_{ij}(a) \lf( {\beta}   \cdot  (  h^t_j - h^t_{i} +r_{ij}(a) -M ) \ri)^2\\
&\leq \sumi \suma  \sumj p_{ij}(a) \lf( {\beta} \cdot  2 \cdot (4\tmix +1)\ri)^2\\
&= { 4\S \A  \beta^2 (4\tmix+1)^2 },
\end{align*}
where the inequality uses the fact that $h^t\in\cH$.
\qed




\begin{lemma}  
The iterates generated by Algorithm 1 satisfy
\begin{equation}\label{eq-Phi}
\E\[ \KL({\mu}^* || {\mu}^{t+1} ) \mid \cF_t\]  \leq \KL({\mu}^* || {\mu}^t)  \\+{\beta} \suma (\bpi_a^t - \bpi_a^*)^{\top} \lf( (   P_a - I ) \bh^t + \br_a\ri) +  2\S \A (4\tmix+1)^2\beta^2,
\end{equation}
for all $t\geq 0$, with probability 1.
\end{lemma} 

\pf 
For arbitrary $i\in\cS$ and $a\in\cA$, we have
$$\frac{1}\beta\cdot \E\[ \g^{t+1}_{{i,a}}  \mid \cF_t\] =   \sumj p_{ij}(a)h^t_j  -h^t_{i} + \sumj p_{ij}(a) r_{ij}(a) -M  = (  P_a \bh^t - \bh^t + \br_a)_i - M.$$
It follows that
\begin{align*}
\frac{1}\beta\cdot \sumia (\bpi^t_{{i,a}}  - \bpi^*_{{i,a}}) \E\[ \g^{t+1}_{{i,a}}  \mid \cF_t\] 
&= \suma\sumi(\bpi^t_{i,a} - \bpi^*_{i,a})\[ 
(  P_a \bh^t - \bh^t + \br_a)_i - M
 \] \\
 &= \suma (\bpi_a^t - \bpi_a^*)^{\top} \lf( (   P_a - I ) \bh^t + \br_a\ri),
\end{align*}
where the second equality comes from the fact $\sumia \bpi^t_{{i,a}} = \sumia  \bpi^*_{{i,a}} =1$ (because $\bpi^t\in\cU$, $\bpi^*\in\cU$).
We further apply Lemmas 2-\ref{lemma-variance} and complete the proof.

\qed



\begin{lemma} \label{lemma-v}
The iterates generated by Algorithm 1 satisfy for all $t\geq 0$ with probability 1 that
\begin{equation}\label{eq-v}
\E\[\|\bh^{t+1}-\bh^*\|^2 \mid \cF_t\]
\leq \|\bh^t-\bh^*\|^2 + 2 {\alpha} (\bh^t-\bh^*)^{\top} 
\lf(\suma  (I-  P_a)^{\top}\bpi_a^t \ri)+ \cO\lf( \alpha^2 \ri).
\end{equation}
\end{lemma} 

\pf 
According to the updates of Algorithm 1, we have
$$\bh^{t+1} = \mathbf{Proj}_{\cH} \[\bh^t +  \bd^t \] ,$$
where $\mathbf{Proj}_{\cH}$ denotes the Euclidean projection onto $\cH = \{\bh \mid \|\bh\|_{\infty}\leq 2\tmix\}$
By using the nonexpansive property of $\Pi_{\cH}$ and $\bh^*\in \cH$, we further obtain
\begin{align*}
\E\[\|\bh^{t+1}-\bh^*\|^2 \mid\cF_t\]
&= \E\[\| \mathbf{Proj}_{\cH}  [\bh^t + \bd^t ] -\bh^*\|^2 \mid\cF_t\]
\leq \E\[\|\bh^t + \bd^t -\bh^*\|^2 \mid\cF_t\]\\
&= \|\bh^t-\bh^*\|^2 +2  (\bh^t-\bh^*)^{\top}\E\[ \bd^t \mid\cF_t\]  +  \E\[\|\bd^t \|^2\mid\cF_t\],
\end{align*}
for all $t$ with probability 1.
We can verify that
$$
\E\[ \bd^t \mid\cF_t\] 
=  \alpha (I-  P^{\pi_t})^{\top}\xi^t  
= \alpha \suma (I-P_a)^{\top} \mathbf{diag}(\pi^t_a) \xi^t
=  \alpha   \suma  (I-  P_a)^{\top}\bpi_a^t .
$$
and
 \begin{align*}
\E\[\bd^t \|^2\mid\cF_t\]  = \cO(\alpha^2).
\end{align*}
Finally we combine all preceding inequalities and obtain \eqref{eq-v}.
\qed


%
%

\begin{lemma} 
We define for short that 
$$\cE^t =  \KL({\mu}^* || {\mu}^t) + \frac{1}{2\S (\tmix)^2} \|\bh^t-\bh^*\|^2
,\quad
\cG^t  
=  \sumia \bpi^t_{{i,a}}( \bh^* - \a P_{a} \bh^* - \br_{a})_i +  \bar v^*. $$
Let $\alpha = \S (\tmix)^2 \beta$.
The iterates generated by Algorithm 1 satisfy for all $t$ with probability 1 that
\begin{equation}\label{thm1:equ5}
\E\[\cE^{t+1}\mid \cF_t\]
\leq \cE^{t} -  \beta \cG^t 
+ {\beta^2}  \tO(\S\A  (\tmix)^2 ).
\end{equation}
\end{lemma} 

\pf Let $\alpha = \S (\tmix)^2 \beta$. We multiply \eqref{eq-v} with $\frac{1}{2\S (\tmix)^2}$ and takes its sum with \eqref{eq-Phi}, obtaining 
\begin{equation*}
\begin{split}
\E\[\cE^{t+1}\mid \cF_t\]
&\leq \cE^{t} + {\beta^2}  \tO(\S \A (\tmix)^2 ) \\
&+ {\beta}  \lf( \suma (\bpi_a^t - \bpi_a^*)^{\top} \lf( (   P_a - I ) \bh^t + \br_a\ri)  + (\bh^t-\bh^*)^{\top} \lf(\suma  (I-  P_a)^{\top}\bpi_a^t  \ri) \ri) .
\end{split}
\end{equation*}
We have
\begin{align*}
& \suma (\bpi_a^t - \bpi_a^*)^{\top} \lf( (   P_a - I ) \bh^t + \br_a\ri)  + (\bh^t-\bh^*)^{\top} \lf(\suma  (I-  P_a)^{\top}\bpi_a^t  \ri)
\\
&= \suma (\bpi_a^t - \bpi_a^*)^{\top} \lf( (   P_a - I ) \bh^t + \br_a\ri)  + (\bh^t-\bh^*)^{\top} \suma  (I-  P_a)^{\top} ( \bpi_a^t -  \bpi_a^*)\quad (\hbox{by the dual feasibility of $\mu^*$}) \\
&= \suma (\bpi_a^t - \bpi_a^*)^{\top} \lf( (   P_a - I ) \bh^* + \br_a\ri) \\
&= \suma (\bpi_a^t)^{\top} \lf( (   P_a - I ) \bh^* + \br_a  \ri)   - \suma \bar v^*\cdot (\bpi_a^*)^{\top}\be
 \quad (\hbox{by the linear complementarity condition for $\bh^*$, $\mu^*$}) \\
 &= \suma (\bpi_a^t)^{\top} \lf( (   P_a - I ) \bh^* + \br_a  \ri)   -\bar v^*
 ,
 \end{align*}
 where the first equality uses the dual feasibility of $\mu^*$ of linear program \eqref{eq-dual}: 
 $$\suma (I-  P_a)^{\top}\bpi^*_a =0,$$ 
 and the third equality uses the complementary condition of the linear programs \eqref{eq-primal}-\eqref{eq-dual}:
$$\bpi_{a,i}^* \lf( (   P_a - I ) \bh^* + \br_a -  \bar v^*\cdot \be \ri)_i =0,\qquad \forall~i\in\cS,a\in\cA.$$
 Combining the preceding relations, we obtain  \eqref{thm1:equ5}. 
\qed

\paragraph{Proof of Theorem \ref{thm-dualitygap}.}
We claim that $\cE^1\leq \log(\S\A) + 2$. To see this, we note that ${\mu}^1$ is the uniform distribution (according to Step 2 of Algorithm 1) and $\bh^0,\bh^*\in\cH$. Therefore we have $\KL({\mu}^*||{\mu}^1)\leq \log (\S\A)$ and $\|\bh^t- \bh^*\|^2\leq  4\S (\tmix)^2$ for all $t$. Then we have $\cE^1 \leq  \KL({\mu}^* || {\mu}^1)   + \frac{1}{2\S (\tmix)^2 } \|\bh^1-\bh^*\|^2 \leq \log(\S\A) + 2$.

We rearrange the terms of \eqref{thm1:equ5} and obtain
$$
 \gk 
      \leq \frac{1}{\beta }(\ek-  \E\[ \cE^{t+1} \mid \cF_t\])+ \beta \tO(\S\A  (\tmix)^2 )   .
$$
Summing over $t=1,\ldots,T$ and taking average, we have 
   \begin{equation*}
       \begin{aligned}
             \E\[  \sum_{t=1}^{T}  \gk \] 
          & \leq \frac{1}{ \beta  }\sum_{t=1}^{T}( \E\[\ek\]- \E\[ \cE^{t+1} \])+T\beta \tO(\S\A  (\tmix)^2 ) \\
           &= \frac{\E\[\cE^1\]-\E\[\cE^t\]}{\beta } +T\beta \tO(\S\A  (\tmix)^2) \\
           &\leq \frac{1}{\beta  } (\log(\S\A)+2)+ T\beta \tO(\S\A  (\tmix)^2 ) .
     \end{aligned}
   \end{equation*}
where the inequality is based on the fact $\cE^1 \leq \log(\S\A) + 2$ and $\cE^t\geq 0$. 
Therefore by taking $\beta  =\frac1{\tmix} \sqrt{\frac{\log \S\A }{2\S \A T} }$,
we obtain
$
  \E\[  \frac{1}{T}\sum_{t=1}^{\top}\gk \]   =\tO\lf( \tmix \sqrt{\frac{\S\A}{T}}\ri).
$
\qed

\end{document}